\documentclass{article}
\usepackage[preprint]{neurips_2026}
\usepackage[utf8]{inputenc}
\usepackage[T1]{fontenc}
\usepackage{hyperref}
\usepackage{url}
\usepackage{booktabs}
\usepackage{amsfonts}
\usepackage{nicefrac}
\usepackage{microtype}
\usepackage{graphicx}
\usepackage{xcolor}
\usepackage{subcaption}
\usepackage{multirow}
\usepackage{pifont}
\usepackage{float}
\usepackage{algorithm}
\usepackage{algpseudocode}
\usepackage{rotating}
\newcommand{\cmark}{\ding{51}}
\newcommand{\xmark}{\ding{55}}

\title{DW-Bench: Benchmarking LLMs on Data Warehouse\\Graph Topology Reasoning}

\author{
  Ahmed G.A.H Ahmed \\
  Innosol / Bah\c{c}e\c{s}ehir University \\
  \texttt{ahmed.gamal@innosol-labs.eu} \\
  \And
  C. Okan Sakar \\
  Department of Computer Engineering, Bahcesehir University, Turkey \\
  \texttt{okan.sakar@bau.edu.tr} \\
}

\begin{document}

\maketitle

\begin{abstract}
Text-to-SQL benchmarks test whether a model can write the right query.
They do not test whether it understands the schema it queries.
A data warehouse schema is a graph: tables linked by foreign keys,
rows anchored to source records through ETL lineage.
Tracing that structure (finding join paths, detecting disconnected silos,
propagating impact through lineage chains) is routine data engineering,
but no existing benchmark measures it.

We present \textbf{DW-Bench}, a benchmark for graph topology reasoning
over data warehouse schemas.  It contains
1,046~questions (13~subtypes, three difficulty levels, five datasets,
262~tables).  We run six baselines,
from flat context injection to tool-calling and code execution,
against Gemini~2.5 Flash, DeepSeek-V3, and Qwen2.5-72B.

Tool-augmented baselines reach 87--90\% micro-EM; static methods
top out at 63--81\%.  The gap narrows on easy subtypes and widens
on hard ones: no method exceeds 61\% while the oracle clears
$\geq$99.5\%.  That gap is a reasoning problem.
Obfuscating table names drops static baselines 9--32~pp; tool-use
loses under 4~pp.

Code, data, and per-question results:
\url{https://github.com/AJamal27891/dw-bench}.
\end{abstract}

\section{Introduction}

Text-to-SQL has made rapid progress: state-of-the-art LLMs now approach or
match human-level accuracy on Spider~\cite{yu2018spider} and
BIRD~\cite{li2023bird}, which measure a model's capacity to translate a
natural-language question into a SQL query.  However, even the latest
benchmarks like Spider~2.0~\cite{lei2024spider2} show that enterprise-scale
SQLworkflows remain largely unsolved (best models~$\sim$17--21\%).

Data warehouses, on the other hand, present a different kind of problem.
A data engineer maintaining an enterprise warehouse does not primarily write
SQL.  Instead, they trace graph structure: lineage edges, foreign key (FK) paths,
disconnected schema silos.  Common questions include:

\begin{itemize}
  \item \emph{``If the \texttt{Person} table changes, which downstream tables
    are affected through data lineage?''} (impact analysis)
  \item \emph{``Which tables form disconnected silos in our schema graph?''}
    (connectivity analysis)
  \item \emph{``What is the shortest FK path between two tables?''}
    (routing)
\end{itemize}

Answering these questions demands breadth-first search (BFS), connected-component detection, and
multi-hop path enumeration.  No existing benchmark assesses LLMs on these capabilities
over real-world data warehouse schemas with heterogeneous
edge semantics.

Our evaluation reveals three principal findings:
(i)~all models achieve near-perfect scores on single-hop
structural queries but drop 30--40~percentage points on
compositional multi-hop tasks, exposing a systematic
\emph{structural reasoning ceiling};
(ii)~Tool-Use prompting closes the gap on topology-enumeration
subtypes but fails on compositional lineage-impact queries
that require chaining multiple graph algorithms; and
(iii)~obfuscating table names degrades performance by up to
32~pp on lineage tasks, confirming that models partially
rely on lexical cues rather than genuine graph traversal.

Our contributions are summarized as follows:

\begin{enumerate}
  \item \textbf{DW-Bench}: 1,046 schema-level questions
    across 5 data warehouse schemas (4 real-world, 1 synthetic), covering
    3 categories (lineage impact, schema routing, silo detection),
    13 subtypes, and 3 difficulty levels.
  \item \textbf{Six baselines} (Flat Text, Vector-RAG, Graph-Augmented,
    Tool-Use, ReAct-Code, Oracle), tested with Gemini~2.5 Flash,
    DeepSeek-V3, and Qwen2.5-72B.
  \item \textbf{An obfuscation protocol} that randomizes table names,
    enabling controlled evaluation of memorization versus genuine topology
    understanding.
  \item \textbf{Syn-Logistics}: A synthetic dataset with $n \geq 20$ per
    subtype.  It exposes a model-dependent structural gap: Gemini scores
    93--100\% on \texttt{hop\_count}/\texttt{count}, while DeepSeek drops to
    27--48\% on the same subtypes.
  \item \textbf{Open-source release} of all datasets, evaluation code, and
    baseline implementations.
\end{enumerate}

\section{Related Work}

\paragraph{Text-to-SQL Benchmarks.}
Spider~\cite{yu2018spider} established the standard benchmark setting, with 10,181 questions spanning 200 databases. BIRD~\cite{li2023bird} introduced noisier and more realistic schemas, while Spider~2.0~\cite{lei2024spider2} scaled the setting toward enterprise-style schemas. ScienceBenchmark focuses on scientific databases. RobuT~\cite{zhao2023robust} evaluates cell-level and row-level reasoning. All of these benchmarks target \emph{query generation or cell-level reasoning}; none explicitly evaluate structural understanding.

\paragraph{Schema Understanding.}
BEAVER~\cite{beaver2024} evaluates LLMs on database knowledge including
schema matching and entity resolution, but does not evaluate graph topology
reasoning.  LLM-FK~\cite{tang2025llmfk} evaluates FK discovery, a single-hop
structural task complementary to DW-Bench's multi-hop topology reasoning.

\paragraph{Graph Reasoning in LLMs.}
NLGraph~\cite{wang2023nlgraph} and GraphQA~\cite{fatemi2023talk} evaluate LLMs
on synthetic graph problems (connectivity, shortest path).
GraCoRe~\cite{gracore2024} taxonomizes graph reasoning;
GraphArena~\cite{grapharena2024} benchmarks graph algorithms at scale.
However, none of these use real database schemas, and none combine FK and lineage edge types.

\paragraph{Graph-Centric Agents and GNN Hybrids.}
Think-on-Graph~\cite{tog2024} interleaves LLM reasoning with knowledge graph exploration;
Graph Chain-of-Thought~\cite{gcot2024} augments prompts with graph-derived
reasoning chains.  G-Retriever~\cite{gretriever2024} uses GNN encoders to
retrieve the graph context for LLM QA.  Retrieval-augmented
generation~\cite{rag2020} (RAG) and hierarchical
GraphRAG~\cite{graphrag2024} address knowledge-intensive tasks but
do not specifically target schema topology. Recent work on tool-based graph
reasoning agents (e.g., GraphWalk) explicitly studies scale and iteration 
budgets for compositional navigation, while structural-reward training frameworks
(e.g., STRuCT-LLM) offer alternative methods to scaffold compositional reasoning.
DW-Bench does not compare against these advanced graph-agent frameworks;
following the precedent of Spider and BIRD, we evaluate standardized,
zero-shot baselines to establish a \emph{lower bound}.  These methods
are natural next steps: the Oracle upper bound ($\geq$99.5\% EM) quantifies
exactly how much room remains for hybrid architectures combining tools with
learned graph representations.

\paragraph{Positioning.}
DW-Bench uniquely combines: (1) \emph{real-world} schemas from industry
standards, (2) \emph{heterogeneous} edge types (FK + lineage), (3)
\emph{practical} question types from data engineering workflows, and (4)
\emph{obfuscation} for contamination control.

\section{DW-Bench: Benchmark Design}

\subsection{Datasets}

\begin{table}[t]
\centering
\caption{DW-Bench dataset statistics. ``Silos'' = disconnected
components in the combined FK+lineage graph.}
\label{tab:datasets}
\small
\begin{tabular}{@{}lcccccc@{}}
\toprule
Dataset & Domain & Tbl & FK & Lin. & Qs & Silos \\
\midrule
AdventureWorks & Retail   & 102 & 136 & 39 & 208 & 11 \\
TPC-DS         & Analytics&  24 &  70 &  0 & 127 &  1 \\
TPC-DI         & ETL      &  35 &  29 & 21 & 181 &  2 \\
OMOP CDM       & Health   &  37 &  74 & 21 & 158 &  3 \\
Syn-Logistics  & Supply   &  64 &  96 & 35 & 372 &  5 \\
\midrule
\textbf{Total} &          & \textbf{262} & \textbf{405} & \textbf{116} & \textbf{1046} & --- \\
\bottomrule
\end{tabular}
\end{table}

We select five datasets shown in Table~\ref{tab:datasets} representing diverse data warehouse topologies:

\textbf{AdventureWorks}~is a Microsoft reference data warehouse with 102
tables spanning OLTP and DW layers, connected by 136 FK edges and 39
lineage (derived\_from) edges. Its dual-layer structure makes it ideal for
lineage impact questions.

\textbf{TPC-DS}~is the industry standard analytics benchmark with a star
schema of 24 tables and 70 FK edges. Its single connected component and
absence of lineage edges tests pure FK-based reasoning.

\textbf{TPC-DI}~is the data integration benchmark with 35 tables modeling
an ETL pipeline from staging to warehouse, with 21 lineage edges
representing the transformation flow.

\textbf{OMOP CDM}~(Observational Medical Outcomes Partnership Common Data
Model)~\cite{ohdsi2019omop} is a healthcare standard with 37 tables, 74
FK edges, and 21 lineage edges across 3 connected components (clinical
data, vocabulary tables, and metadata).

\textbf{Syn-Logistics}~is our own synthetic supply-chain schema: 64
tables, 5 connected components (carrier, procurement, healthcare, finance,
HR).  We built it to fix a statistical-power problem in the real-world
corpus: every subtype gets $n \geq 20$ questions.  Table names come from
a domain dictionary so obfuscation does not collapse semantics.

\subsection{Schema Graph Representation}

Each dataset is represented as a heterogeneous graph $G = (V, E_{fk} \cup
E_{lin})$, where $V$ denotes the set of table nodes, $E_{fk}$ denotes the
set of foreign-key edges, and $E_{lin}$ denotes the set of data-lineage
edges, using PyTorch Geometric~\cite{fey2019pyg}
\texttt{HeteroData}. Each node represents a table and is associated with six structural features: in-degree, out-degree, normalized
degree, lineage degree, betweenness centrality, and PageRank. Edges are
typed as either \texttt{fk\_to} (foreign key) or \texttt{derived\_from}
(data lineage). Lineage edges form a strict DAG (directed acyclic graph):
each \texttt{derived\_from} edge points from a downstream DW table to its
upstream source, with no cycles or self-references. All traversal
algorithms respect edge typing: FK queries use only \texttt{fk\_to}
edges, lineage queries use only \texttt{derived\_from}, and
\texttt{combined\_impact} queries compose both types sequentially.
FK edges are treated as \textbf{directed} (parent$\to$child) for
path and hop-count queries (\texttt{join\_path}, \texttt{hop\_count},
\texttt{direct\_fk}) and as \textbf{undirected} for connectivity
queries (\texttt{membership}, \texttt{connected}, \texttt{isolation},
\texttt{count}, \texttt{full\_enum}).Lineage edges are always directed (downstream$\to$upstream),
following the \texttt{derived\_from} convention defined above.
Foreign keys are treated as bidirectional (undirected) explicitly for
connectivity and silo-detection subtypes, as an FK implies an inherent
structural relationship that can be traversed logically in either direction
(e.g., from a parent lookup table to child fact rows, or vice versa).

For path-based questions, we validate alternative valid shortest paths
against the FK adjacency matrix, ensuring EM is robust to tie-breaking.

\subsection{Question Generation}
\label{sec:qa-generation}

Questions are generated deterministically from the graph structure using
standard graph algorithms (BFS, connected component detection)
implemented with NetworkX~\cite{networkx2008}, ensuring ground-truth
answers are provably correct. We define three categories with 13 subtypes as shown in Table~\ref{tab:taxonomy}. 

\paragraph{Difficulty assignment.}
Difficulty labels are assigned a priori based on structural complexity,
following the precedent of Spider~\cite{yu2018spider} (which uses SQL AST
depth) and BIRD~\cite{li2023bird}.  Easy subtypes require a single graph
lookup (one hop, one edge type).  Medium subtypes require multi-hop
traversal within a single edge type.  Hard subtypes require either
multi-hop transitive closure or composition across both FK and lineage
edge types.  These labels are fixed properties of the question structure,
independent of any model's empirical performance.

\begin{table}[t]
\centering
\caption{Question taxonomy. Difficulty is assigned based on the number of
reasoning hops and edge types required.}
\label{tab:taxonomy}
\small
\begin{tabular}{@{}llclr@{}}
\toprule
Category & Subtype & Diff. & Description & \#Qs \\
\midrule
\multirow{5}{*}{\rotatebox{90}{\small Lin.}} 
  & forward         & Easy & Direct lineage targets   & 72 \\
  & reverse         & Easy & Source tables for DW table & 71 \\
  & multi\_source   & Med. & Tables with 3+ sources   & 49 \\
  & transitive      & Hard & Multi-hop lineage chains  & 33 \\
  & combined\_impact & Hard & Lineage + FK dependents  & 69 \\
\midrule
\multirow{3}{*}{\rotatebox{90}{\small Rt.}}
  & direct\_fk      & Easy & FK adjacency check       & 100 \\
  & join\_path      & Med. & FK shortest path         & 342 \\
  & hop\_count      & Med. & Path length              & 124 \\
\midrule
\multirow{5}{*}{\rotatebox{90}{\small Silo}}
  & count           & Easy & Number of components     & 30 \\
  & isolation       & Med. & Is table X isolated?     & 60 \\
  & connected       & Med. & Are X and Y connected?   & 29 \\
  & membership      & Hard & Which component has X?   & 38 \\
  & full\_enum.     & Hard & List all tables in silo  & 29 \\
\bottomrule
\end{tabular}
\end{table}

\subsection{Obfuscation Protocol}

To distinguish genuine topology-based reasoning from reliance on
surface lexical cues and schema-name memorization,
we generate an \emph{obfuscated} variant for each dataset.  Table names
are replaced with random identifiers (\texttt{Table\_A},
\texttt{Table\_B}, etc.) using a deterministic mapping.
This prevents models from exploiting semantic signals in
names such as \texttt{Customer} or \texttt{Invoice} to infer
relationships without genuine structural reasoning.
Questions and answers are updated accordingly with
word-boundary-aware replacement to avoid corrupting natural
language phrases (e.g., the word ``relationship'' in
``foreign key relationship'' is preserved).

\section{Baselines}
\label{sec:baselines}

Six baselines are evaluated under a shared prompt with different context injection strategies, all in the zero-shot setting. Each paradigm receives the form of context that is natural to its architecture, rather than being constrained to token-budget parity, since the goal is to compare \emph{reasoning paradigms} rather than token counts. The baseline methods are briefly summarized below.

\textbf{Flat Text (FT).}  Dump the whole schema (tables, columns, FKs,
lineage) as plain text.  This mirrors what a developer does when pasting
data definition language (DDL) into a long-context model.

\textbf{Vector-RAG (VR).}  Embed schema elements with
Sentence-BERT~\cite{reimers2019sentencebert}, retrieve $k{=}15$ chunks via
FAISS~\cite{faiss2019}, inject as context.

\textbf{Graph-Augmented (GA).}  Extract a 3-hop BFS neighborhood around the
tables mentioned in the question.  For global questions with no anchor
tables (``how many components?''), supply the full graph instead.

\textbf{Tool-Use (TU).}  Provide the model with nine graph reasoning tools:
\texttt{shortest\_path}, \texttt{connected\_\allowbreak{}components},
\texttt{get\_component\_of}, \texttt{get\_fk\_\allowbreak{}neighbors},
\texttt{get\_lineage\_forward}, \texttt{get\_lineage\_\allowbreak{}reverse},
\texttt{transitive\_lineage}, \texttt{check\_fk\_\allowbreak{}adjacency},
\texttt{list\_tables}. A maximum of three tool calls is allowed per question.

\textbf{ReAct-Code (RC).}  The model writes and runs Python/NetworkX code
in a sandboxed REPL, up to 5 rounds.  More flexible than TU and serves as
a budget ablation (5 code rounds vs.\ TU's 3 tool calls).

\textbf{Oracle.}  Gold algorithmic outputs are injected into the prompt.  This
establishes the upper bound.

\section{Experiments}

\subsection{Setup}

We evaluate with \textbf{Gemini~2.5 Flash}~\cite{gemini2024} (closed
frontier), \textbf{DeepSeek-V3}~\cite{deepseekv3} (671B MoE, open-weight),
and \textbf{Qwen2.5-72B}~\cite{qwen25} (72B dense, open-weight),
using greedy decoding ($\tau{=}0$) for deterministic reproducibility.
We report pooled \textbf{Micro-EM} (overall exact match) and
\textbf{Macro-EM} (mean of per-subtype EM, weighting all 13~subtypes
equally).  For list-typed topology tasks (e.g., connected component
membership), we apply \textbf{target-node normalization} (stripping the
queried table from both prediction and gold prior to scoring) ensuring
models are not penalized for harmless self-inclusion.
An \textbf{Oracle} baseline injects gold algorithmic outputs, establishing
an upper bound ($\geq$99.5\% EM across all three models; the residual gap
reflects minor formatting, not reasoning failures).

\subsection{Main Results}

Results are shown in Table \ref{tab:main-results} and the main findings are summarized below:

\begin{table}[t]
\centering
\caption{Main results (EM~\% $\pm$ 95\% CI, pooled across
5 datasets, 2000 bootstrap resamples).
\textbf{Bold}~=~best non-oracle.}
\label{tab:main-results}
\small
\begin{tabular}{@{}llccccc@{}}
\toprule
Model & & FT & VR & GA & TU & RC \\
\midrule
\multirow{2}{*}{Gem.}
  & Micro & 76.6 & 72.9
  & 75.4 & \textbf{89.3}
  & 81.4 \\
  & Macro & 77.0 & 72.4
  & 77.1 & \textbf{86.6}
  & 69.9 \\
\midrule
\multirow{2}{*}{DS}
  & Micro & 69.5 & 71.0
  & 76.8 & \textbf{90.4}
  & 79.4 \\
  & Macro & 68.2 & 62.4
  & 69.5 & \textbf{88.4}
  & 71.6 \\
\midrule
\multirow{2}{*}{Qw.}
  & Micro & 63.2 & 69.5
  & 80.9 & \textbf{87.5}
  & 64.6 \\
  & Macro & 64.8 & 62.8
  & 77.5 & \textbf{82.5}
  & 62.2 \\
\midrule
\multicolumn{2}{l}{\textit{Oracle}}
  & \multicolumn{5}{c}{99.5 / 99.8 / 100.0} \\
\bottomrule
\end{tabular}
\end{table}

\paragraph{Finding~1: Agentic baselines dominate.}
TU tops the ranking at 87--90\% micro-EM, roughly 7--14~pp ahead of the best
static baseline.  RC also clears static methods through code generation.

\paragraph{Finding~2: Triviality Illusion.}
FT looks competitive at 76.6\% micro-EM, but the illusion comes from
\texttt{join\_path} (33\% of questions) inflating the average.
Macro-EM, which weights subtypes equally, puts GA ahead of FT among
static methods; all still trail TU by $>$8~pp (Appendix Figure~\ref{fig:triviality}).

\subsection{Results by Difficulty}

\begin{table}[t]
\centering
\caption{EM (\%) by difficulty. TU achieves near-perfect easy scores
but plateaus on hard questions alongside static baselines.}
\label{tab:difficulty}
\small
\begin{tabular}{@{}llccccc@{}}
\toprule
Model & Diff. & FT & VR & GA & TU & RC \\
\midrule
\multirow{3}{*}{Gem.}
  & Easy    & 82.7 & 82.5 & 82.2 & \textbf{99.6} & 90.4 \\
  & Med.    & 78.9 & 73.0 & 76.3 & \textbf{90.3} & 78.7 \\
  & Hard    & 56.1 & 46.5 & 55.6 & 59.9 & \textbf{60.9} \\
\midrule
\multirow{3}{*}{DS}
  & Easy    & 82.7 & 84.4 & 87.1 & \textbf{98.0} & 88.8 \\
  & Med.    & 65.3 & 65.3 & 76.0 & \textbf{97.0} & 76.3 \\
  & Hard    & 40.1 & 43.6 & 50.0 & \textbf{60.4} & 58.9 \\
\midrule
\multirow{3}{*}{Qw.}
  & Easy    & 77.4 & 84.0 & 86.6 & \textbf{96.1} & 72.2 \\
  & Med.    & 57.7 & 63.0 & 90.7 & \textbf{92.0} & 64.7 \\
  & Hard    & 33.2 & 40.1 & 51.0 & \textbf{57.4} & 44.1 \\
\midrule
\multicolumn{2}{l}{\textit{Oracle Hard}}
  & \multicolumn{5}{c}{99.0 / 99.5 / 100.0} \\
\bottomrule
\end{tabular}
\end{table}

\paragraph{Easy/Medium patterns.}
As seen in Table \ref{tab:difficulty}, Gemini FT scores 78.9\% on Medium versus 82.7\% on Easy.  Medium is
dominated by \texttt{join\_path} (342 of 544 Medium questions), which has
high FT accuracy since short FK paths are often present in the flat
schema text.  Easy includes subtypes like \texttt{count} (number of
connected components) that require global graph access unavailable to FT.
The distribution reflects subtype composition, not a flaw in difficulty
labels.

\subsection{Per-Subtype Analysis}

\begin{figure*}[t]
\centering
\includegraphics[width=0.85\textwidth]{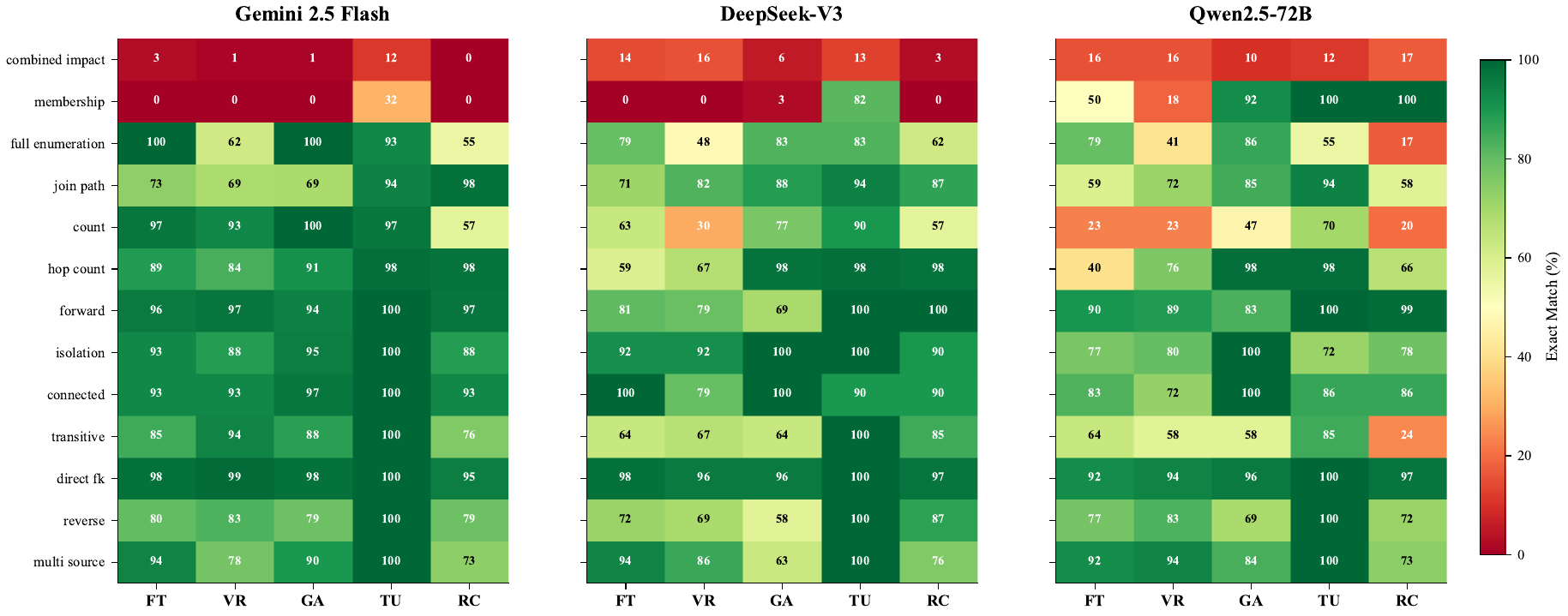}
\caption{EM~(\%) by subtype for all three models.  With
target-node normalization, TU now averages 71\% on
\texttt{membership} (with Qwen reaching 100\%), indicating that topology enumeration is fully
solvable with tools.  \texttt{combined\_impact} remains at 0--17\%,
isolating compositional multi-hop reasoning as the core bottleneck.
\texttt{full\_enumeration} scores reach up to 100\% (Gemini) for FT/GA because these
questions are answered from the full schema context that Flat Text
and Graph-Augmented inject by default for global graph queries.}
\label{fig:heatmap}
\end{figure*}

\paragraph{Finding~3: Hard-task ceiling.}
The results based on subtype analysis is shown in Figure \ref{fig:heatmap}. All baselines plateau at $\sim$60\% on hard questions while Oracle
achieves $>$98\%.  The \texttt{combined\_impact} subtype ($n=69$)
averages only 12\% EM even with TU, identifying compositional
graph reasoning as the fundamental bottleneck.
RC (5~code rounds) shows a similar ceiling to TU (3~tool calls),
suggesting the bottleneck is compositional reasoning itself rather
than interaction budget.

\subsection{Obfuscation}

\begin{table}[t]
\centering
\caption{Obfuscation penalty (4 datasets). TU is nearly invariant.}
\label{tab:obfuscation}
\small
\begin{tabular}{@{}llccccc@{}}
\toprule
Model & & FT & VR & GA & TU & RC \\
\midrule
\multirow{2}{*}{Gem.}
  & Orig.   & 71.8 & 69.3 & 71.2 & \textbf{91.2} & 87.1 \\
  & $\Delta$ & $-$15.0 & $-$25.8 & $-$27.7 & $-$\textbf{3.4} & $-$7.6 \\
\midrule
\multirow{2}{*}{DS}
  & Orig.   & 68.4 & 74.5 & 82.3 & \textbf{91.4} & 83.4 \\
  & $\Delta$ & $-$27.9 & $-$31.9 & $-$24.6 & $-$\textbf{4.2} & $+$0.6 \\
\midrule
\multirow{2}{*}{Qw.}
  & Orig.   & 59.6 & 70.5 & 82.8 & \textbf{88.4} & 62.8 \\
  & $\Delta$ & $-$8.9 & $-$13.8 & $-$19.7 & $-$\textbf{3.7} & $-$7.5 \\
\bottomrule
\end{tabular}
\end{table}

As seen in Table \ref{tab:obfuscation}, TU shows minimal degradation ($\Delta{=}{-}3.4$\% Gemini, ${-}4.2$\%
DeepSeek, ${-}3.7$\% Qwen) while static baselines lose 9--32pp.
Obfuscation replaces table names only; column names are preserved,
as all baselines operate on schema-level graph topology where
column semantics play no role.
RC shows near-zero penalty for DeepSeek ($\Delta{=}{+}0.6$\%), confirming
that code execution against the graph object is largely invariant to
name perturbation.

\subsection{Discussion}

\begin{figure}[t]
\centering
\includegraphics[width=\linewidth]{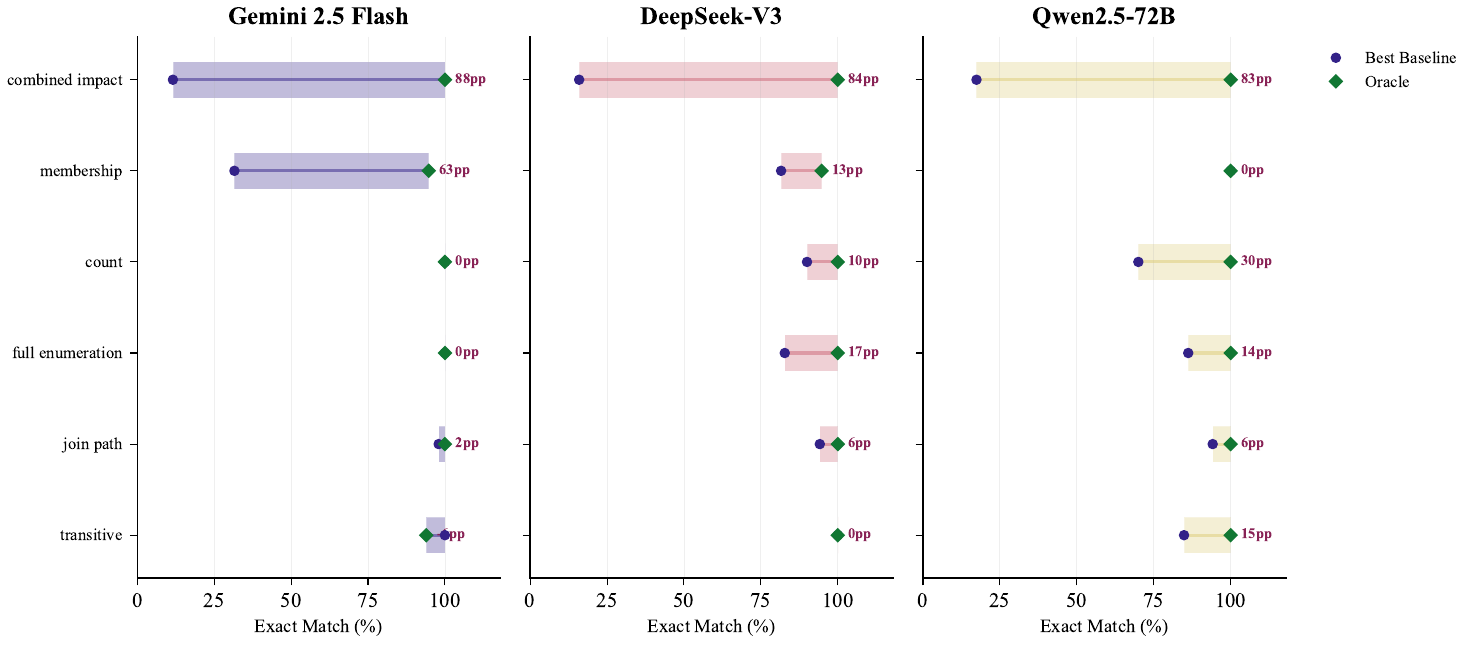}
\caption{Unsolved subtypes: Oracle~EM minus best non-oracle
baseline.  \texttt{combined\_impact} retains $>$82~pp gap;
\texttt{count}, \texttt{full\_enumeration}, and \texttt{transitive}
show 5--25~pp gaps.  The remaining 9 of 13 subtypes are solved
($<$5~pp).}
\label{fig:oracle-gap}
\end{figure}

Our results decompose graph reasoning into \textbf{retrieval-limited}
subtypes (solved by tools) and \textbf{reasoning-limited} subtypes
(unsolved even with tools). The $\sim$35\% of questions that remain
unsolved by both TU and GA largely overlap: their union improves EM by
only 6pp as seen in Figure \ref{fig:oracle-gap}, suggesting the bottleneck is compositional structural reasoning
itself---the capability that GNN encoders are designed to provide.

\textbf{\texttt{combined\_impact} specification.}
Impact is the union of lineage-downstream tables and their FK-reachable
neighbors: lineage transitive closure then FK BFS.  This two-phase
composition drives the 12\% TU ceiling.
RC receives five code-execution rounds
(compared with three tool calls for TU), yet Gemini still scores 0\%
on this subtype. These findings suggest that the 
tool budget is not the primary
bottleneck. The main difficulty appears to lie in
compositional planning.

\textbf{GA obfuscation penalty.}
The large GA $\Delta$ ($-$24 to $-$28~pp) likely reflects two compounding
factors: (1)~loss of semantic cues
(\texttt{DimCustomer}$\to$\texttt{Table\_K}) and (2)~45\% of questions
exceed the 3-hop retrieval radius. Obfuscation
therefore removes the schema-name cues that otherwise help LLMs infer
structure beyond the retrieved fragment.

\begin{sloppypar}
\textbf{Tool tautology.}
TU achieving near-perfect scores on easy subtypes is by design: it
isolates hard failures as genuine reasoning problems.  After
target-node normalization, TU reaches 100\% on \texttt{membership}
(largely due to Qwen's label-agnostic matching) but still
only 12--13\% on \texttt{combined\_impact}.  The limiting
factor therefore does not appear to be tool availability, but
compositional multi-step reasoning.
\end{sloppypar}

\textbf{Graph baselines.}
While our evaluation covers standard retrieval and tool-use paradigms, we
do not evaluate graph-specific agent frameworks (e.g., G-Retriever,
GraphRAG, Think-on-Graph) because they typically require task-specific
training data or finetuning, which violates our strict zero-shot
evaluation protocol. DW-Bench establishes a standardized foundational
layer for future evaluation of these hybrid graph-reasoning architectures.

\subsection{Limitations}

We evaluate three frontier LLMs with deterministic single-pass evaluation;
broader model coverage (e.g., reasoning-optimized variants) would
strengthen generality claims. Furthermore, our obfuscation protocol modifies
only table names; restricting semantic leakage through column names or
standardized suffixes (e.g., \texttt{\_id}) is a topic for future iterations.
Lastly, questions currently follow deterministic templates. While this isolates
structural reasoning, future work should evaluate linguistic robustness
by incorporating paraphrase variations. Additionally, we establish fixed
iteration budgets (e.g., 3 calls for TU, 5 rounds for RC) and retrieval radii (3-hop)
to establish baselines; extensive parametric scaling of inference compute is a topic
for future study. Finally, while our lineage edges are derived from strict documentation
standards, exploring LLM robustness against ambiguous or conflicting ETL provenance
remains an open challenge.

\paragraph{Responsible use.}
This benchmark is intended for scientific evaluation of LLM capabilities
and limitations.  Researchers should not apply automated agentic tools
or sandboxed code execution to unauthorized or production environments
without appropriate safeguards.

\paragraph{Dataset licenses.}
TPC-DS and TPC-DI are available under the TPC EULA (free for research).
OMOP CDM uses the Apache~2.0 license. AdventureWorks is released under
the Microsoft Public License. Syn-Logistics is our creation, released
under MIT.

\section{Conclusion}

DW-Bench tests whether LLMs can perform graph-topology reasoning over data
warehouse schemas. We evaluate six baselines across five datasets with three LLMs, yielding 1,046 schema-level questions.
Four findings are particularly notable.

\textbf{(1)~Tool-use dominance.}  Give the LLM deterministic graph tools
and it reaches 87--90\% micro-EM, beating the best static baseline by
7--14~pp. This indicates that LLMs are better at
orchestrating graph algorithms through tool use than at performing the
same reasoning internally.

\textbf{(2)~Triviality illusion.}  Easy lookup subtypes inflate micro-EM.
Switch to macro-EM (equal subtype weight) and the ranking shifts: what
looked like widespread success reveals significant performance disparities.

\textbf{(3)~Hard-task ceiling.}  No baseline, including agentic
code-generation, exceeds 61\% on hard questions; the Oracle hits $>$99\%.
One subtype, \texttt{combined\_impact} (12\% EM with tools), accounts for
most of the gap: it requires chaining lineage traversal with FK expansion,
a compositional step that current baselines handle poorly.

\textbf{(4)~Obfuscation invariance.} Tool-Use is largely unaffected by
obfuscation ($\Delta \leq 4$~pp), while static baselines decline by
9--32~pp. This result is consistent with tool-based reasoning over graph
topology rather than reliance on table-name cues.

The natural next step is a \textbf{hybrid architecture}: pair interactive
tools with learned graph representations (GNN encoders, graph-aware
controllers) to close the compositional reasoning gap.  DW-Bench supplies
the per-subtype diagnostics needed to measure progress.

\bibliography{references}
\bibliographystyle{abbrvnat}

\appendix

\section{Supplementary Figures}
\label{app:figures}

\begin{figure}[H]
\centering
\includegraphics[width=\linewidth]{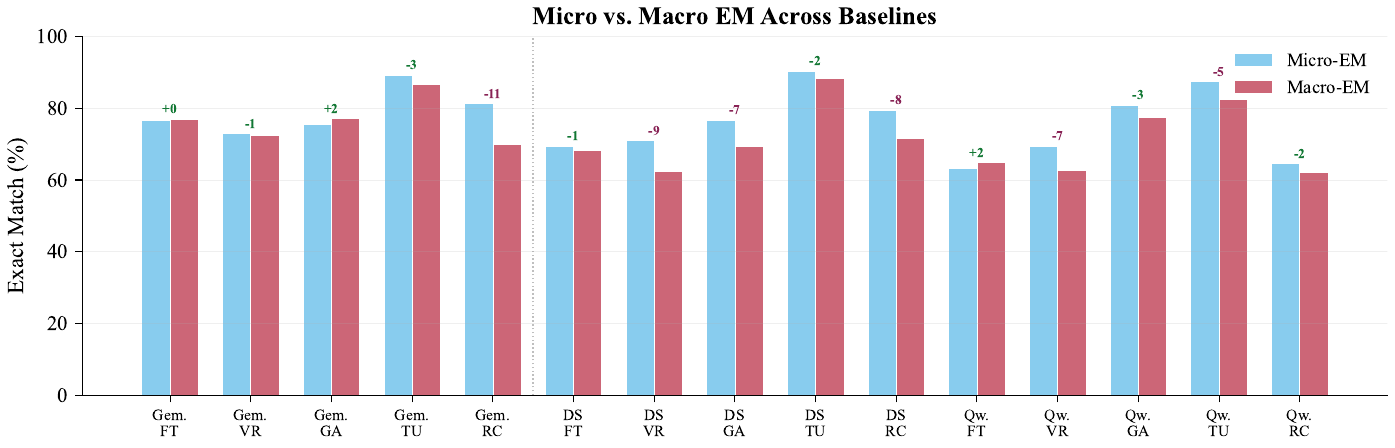}
\caption{The Triviality Illusion: Micro vs.\ Macro EM. Negative deltas
reveal that easy subtypes inflate aggregate scores for static baselines.}
\label{fig:triviality}
\end{figure}

\begin{figure}[H]
\centering
\includegraphics[width=\linewidth]{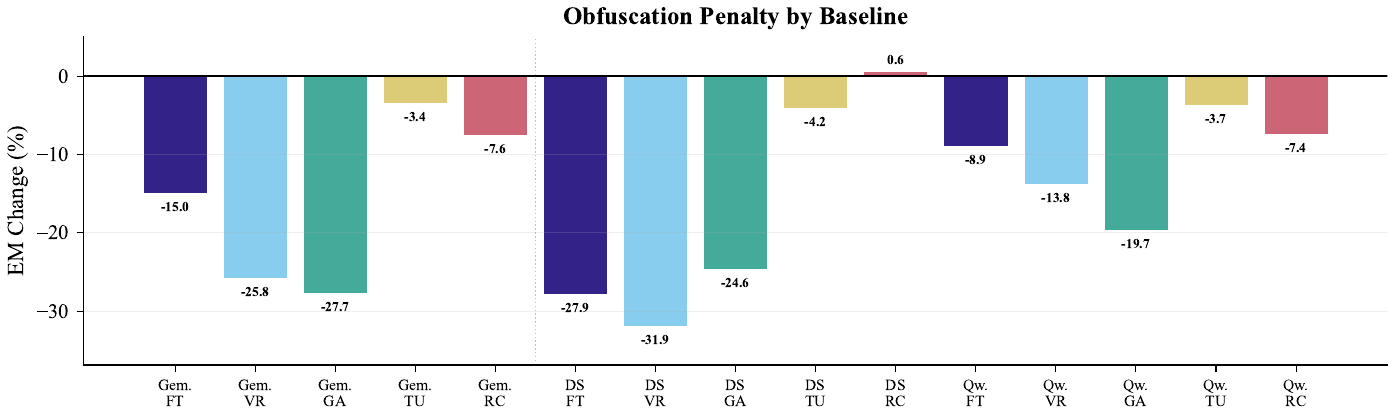}
\caption{Obfuscation penalty by baseline and model. Tool-Use loses
only 3--4\% while static baselines lose 9--32\%.}
\label{fig:obfuscation}
\end{figure}

\section{\texttt{combined\_impact}: Formal Specification}
\label{app:combined-impact}

The \texttt{combined\_impact} subtype captures the set of tables
transitively affected when a source table is modified, propagating
first through lineage then through FK dependencies.

\begin{algorithm}[H]
\caption{\texttt{combined\_impact}($G$, $t$)}
\label{alg:combined-impact}
\begin{algorithmic}[1]
\Require Hetero-graph $G=(V,E_{fk}\cup E_{lin})$; source table $t\in V$
\Ensure Set of affected tables $A\subseteq V$
\State $L \gets \textsc{LineageClosure}(G, t)$
  \Comment{Directed BFS over $E_{lin}$ from $t$}
\State $A \gets L$
\For{each $u \in L$}
  \State $F_u \gets \textsc{FkBFS}(G, u)$
    \Comment{Undirected BFS over $E_{fk}$ from $u$}
  \State $A \gets A \cup F_u$
\EndFor
\State $A \gets A \setminus \{t\}$ \Comment{Exclude source table itself}
\State \Return $A$
\end{algorithmic}
\end{algorithm}

\noindent
\textbf{Corner cases.} (1)~Lineage DAGs contain no cycles by
construction; future datasets with relaxed constraints should detect
and skip back-edges.  (2)~Duplicates during union are naturally
eliminated by set semantics.  (3)~FK BFS treats edges as undirected
(bidirectional), consistent with the connectivity definition for all
Silo-category questions.

\section{Hop-Distance Analysis}
\label{app:hop-distance}

\begin{figure}[H]
\centering
\includegraphics[width=0.85\linewidth]{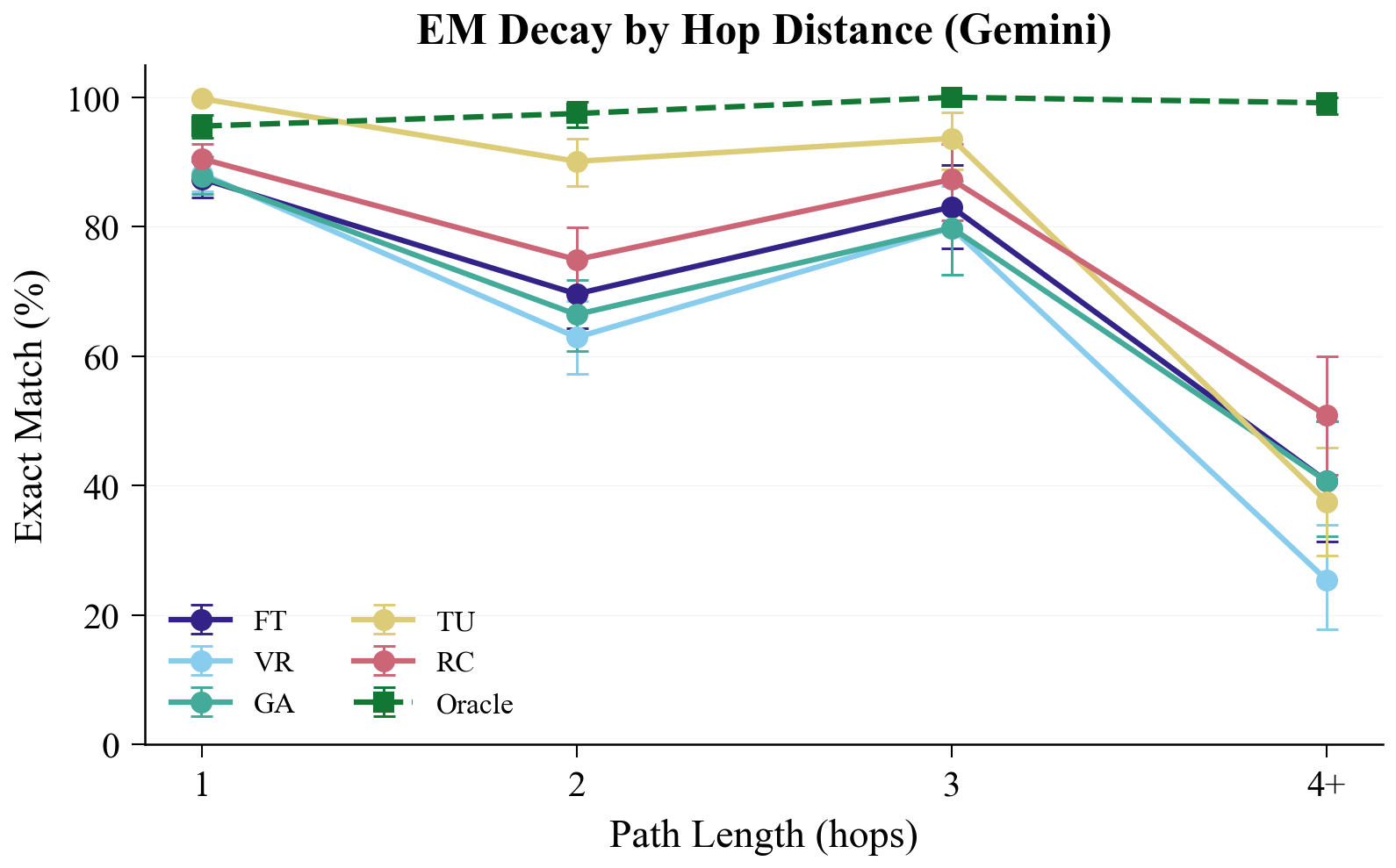}
\caption{EM by gold path length. Static baselines collapse
beyond 3~hops; Tool-Use degrades more gracefully.}
\label{fig:hop_decay}
\end{figure}

\begin{table}[H]
\centering
\caption{Minimum evidence hops per subtype (Syn-Logistics).
45.4\% of questions exceed the GA 3-hop BFS radius.}
\label{tab:hop-distance}
\small
\begin{tabular}{lrrrc}
\toprule
Subtype & $n$ & Exceeds 3h & \% & GA EM \\
\midrule
count            & 26 & 26 & 100\% & 100\% \\
membership       & 20 & 20 & 100\% & 0\% \\
combined\_impact & 34 & 34 & 100\% & 0\% \\
full\_enum       & 25 & 25 & 100\% & 100\% \\
connected        & 25 & 25 & 100\% & 100\% \\
isolation        & 25 & 25 & 100\% & 100\% \\
transitive       & 29 &  4 &  14\% & 93\% \\
hop\_count       & 44 &  5 &  11\% & 100\% \\
direct\_fk       & 36 &  3 &   8\% & 100\% \\
join\_path       & 26 &  2 &   8\% & 81\% \\
multi\_source    & 42 &  0 &   0\% & 95\% \\
forward          & 20 &  0 &   0\% & 100\% \\
reverse          & 20 &  0 &   0\% & 100\% \\
\midrule
\textbf{Total}   & 372 & 169 & \textbf{45.4\%} & 83.1\% \\
\bottomrule
\end{tabular}
\end{table}

\section{Dataset Details}
\label{app:datasets}

\paragraph{Syn-Logistics Design.}
The Syn-Logistics dataset guarantees $n\geq20$ per subtype via three
parallel 4-hop lineage chains across five components. Table names are drawn
from realistic supply-chain terminology (64 tables, 96 FK, 35 lineage).

\paragraph{Lineage Edge Curation.}
\textbf{AW}: Microsoft reference docs (39 edges).
\textbf{TPC-DS}: No lineage (pure FK).
\textbf{TPC-DI}: Official spec ETL flows (21 edges).
\textbf{OMOP}: OHDSI ETL conventions (21 edges).
\textbf{Syn-Log}: By construction (35 edges).

\section{Evaluation Details}
\label{app:decoding}

\paragraph{Models.}
\textbf{Gemini~2.5~Flash} (Google, closed-weight, accessed via API) and
\textbf{DeepSeek-V3} (DeepSeek, 671B MoE, MIT licensed, accessed via API).
Both use greedy decoding ($\tau{=}0$, $p{=}1.0$) with a single
deterministic pass per question for exact reproducibility.

\paragraph{Scoring.}
\emph{Exact Match (EM)}: predicted answer must exactly match the gold
answer after canonicalization.
\emph{Path validation}: for \texttt{join\_path}, alternative shortest
paths of equal length are validated against the FK adjacency matrix.
\emph{List canonicalization}: lists are lowercased and sorted
alphabetically before comparison; order does not matter.
\emph{F1}: token-level overlap between predicted and gold answer sets.

\begin{sloppypar}
\paragraph{Retrieval.}
Vector-RAG uses \texttt{all-MiniLM-L6-v2}~\cite{reimers2019sentencebert}
embeddings with FAISS exact search and $k{=}15$ retrieved chunks.
Graph-Augmented uses 3-hop BFS neighborhoods from mentioned tables.
Tool-Use provides 9 graph algorithm tools
(\texttt{shortest\_path},
\texttt{connected\_components},
\texttt{get\_component\_of},
\texttt{get\_fk\_neighbors},
\texttt{get\_lineage\_forward},
\texttt{get\_lineage\_reverse},
\texttt{transitive\_lineage},
\texttt{check\_fk\_adjacency},
\texttt{list\_tables})
with up to 3 calls per question.
ReAct-Code provides a Python REPL with NetworkX and allows up to 5 code
execution rounds per question.
\end{sloppypar}

\paragraph{Statistical treatment.}
All figures include 95\% bootstrap confidence intervals computed via
2,000 resamples over questions. Per-question results (JSON) are released
for community bootstrap and stratification analyses.

\section{Example Questions and Predictions}
\label{app:examples}

\begin{table}[H]
\centering
\caption{Representative questions across difficulty levels and
categories, with gold answers and Tool-Use (TU) predictions.
\cmark~=~correct, \xmark~=~incorrect.
Tool-Use succeeds on lookup-style questions (direct FK, shortest path)
but fails on exhaustive enumeration (membership) and
multi-hop composition (combined impact).}
\label{tab:examples}
\small
\begin{tabular}{@{}p{0.10\linewidth}p{0.35\linewidth}p{0.24\linewidth}p{0.22\linewidth}@{}}
\toprule
\textbf{Subtype} & \textbf{Question} & \textbf{Gold} & \textbf{TU Pred.} \\
\midrule
\texttt{direct\_fk} \newline \textit{Easy} &
Is there a direct FK between \texttt{DimProduct} and
\texttt{FactInternetSales}? &
yes &
yes \cmark \\
\midrule
\texttt{join\_path} \newline \textit{Medium} &
Shortest FK path from \texttt{DimCustomer} to
\texttt{DimProduct}? &
{[}DimCustomer, FactInternet\-Sales, DimProduct{]} &
{[}DimCustomer, FactInternet\-Sales, DimProduct{]} \cmark \\
\midrule
\texttt{membership} \newline \textit{Hard} &
Which component contains \texttt{DimCurrency}? List all
tables. &
{[}DimCurrency, FactCurrency\-Rate, DimDate, \ldots{]} (12) &
{[}DimCurrency{]} \xmark \\
\midrule
\texttt{combined\_\newline impact} \newline \textit{Hard} &
Which tables are transitively affected (lineage+FK) if
\texttt{raw\_purchase\_orders} is modified? &
{[}stg\_purchase\_orders, dim\_supplier, \ldots{]} (8) &
{[}stg\_purchase\_orders{]} \xmark \\
\bottomrule
\end{tabular}
\end{table}

\section{Partial Correctness (F1 Scores)}
\label{app:f1-scores}

For list-typed target answers (e.g., set enumeration tasks), Exact Match (EM) acts as a strict step-function. To better reflect partial correctness, we report the token-level F1 scores for the three primary enumeration subtypes: \texttt{membership}, \texttt{full\_enumeration}, and \texttt{combined\_impact} (Table~\ref{tab:f1-scores}). F1 scores demonstrate that models retrieve partially correct answers even when failing EM, though the compositional gap remains severe on \texttt{combined\_impact} (F1 peaks at 45.4\% vs.\ 100\% for Oracle).

\begin{table}[H]
\centering
\caption{F1 scores (\%) for enumeration subtypes across models and baselines.}
\label{tab:f1-scores}
\small
\begin{tabular}{@{}llccccc@{}}
\toprule
Subtype & Model & FT & VR & GA & TU & RC \\
\midrule
\multirow{3}{*}{\texttt{membership}} 
  & Gem. & 0.0 & 0.0 & 0.0 & 58.7 & 0.0 \\
  & DS   & 0.0 & 0.0 & 9.4 & 90.1 & 0.0 \\
  & Qw.  & 75.0 & 53.6 & 94.6 & \textbf{100.0} & \textbf{100.0} \\
\midrule
\multirow{3}{*}{\texttt{full\_enum}} 
  & Gem. & \textbf{100.0} & 79.4 & \textbf{100.0} & 94.7 & 69.8 \\
  & DS   & 90.5 & 70.0 & \textbf{91.2} & 90.4 & 77.2 \\
  & Qw.  & 88.6 & 67.8 & \textbf{92.5} & 66.8 & 39.5 \\
\midrule
\multirow{3}{*}{\texttt{combined\_impact}} 
  & Gem. & 17.6 & 11.2 & 11.1 & \textbf{32.9} & 4.3 \\
  & DS   & 23.9 & 22.1 & 14.2 & \textbf{31.5} & 9.1 \\
  & Qw.  & 20.8 & 19.3 & 15.0 & 44.8 & \textbf{45.4} \\
\bottomrule
\end{tabular}
\end{table}

\end{document}